\documentclass[journal]{IEEEtran}

\usepackage{amsmath,amsfonts}
\usepackage{algorithmic}
\usepackage{subcaption}
\usepackage{booktabs}
\usepackage{array}
\usepackage{textcomp}
\usepackage{stfloats}
\usepackage{url}
\usepackage{verbatim}
\usepackage{graphicx}
\usepackage{xcolor}
\usepackage[table]{xcolor}
\usepackage{multirow}
\usepackage{makecell}
\usepackage{hyperref}
\ifCLASSINFOpdf
\else
\fi
\begin{document}
%\title{Standardized Evaluation of Light Field Quality Metrics under Coding and View Synthesis Distortions}

%\title{Standardized Evaluation of Objective Light Field Quality Metrics: Benchmark Design, Hybrid Assessment, and Anchor Analysis}

\title{Towards Standardized Light Field Quality Assessment: Hybrid Subjective Benchmarking and Objective Metric Evaluation}


%\author{Michael~Shell,~\IEEEmembership{Member,~IEEE,}
%        John~Doe,~\IEEEmembership{Fellow,~OSA,}
%        and~Jane~Doe,~\IEEEmembership{Life~Fellow,~IEEE}% <-this % stops a space

\author{Saeed Mahmoudpour, Myl\`ene C. Q. Farias, Gi-Mun Um, Myllena A. Prado, Ismael Seidel, Leonardo de Sousa Marques, Leonardo Andrade, Shengyang Zhao, Carla L Pagliari,~\IEEEmembership{Senior Member,~IEEE} 

\thanks{This work is funded by Electronics and Telecommunications Research Institute under Collaboration Agreement - 86203.}
\thanks{Saeed Mahmoudpour is with the Dept. of Electronics and Informatics, Vrije Universiteit Brussel, Belgium and with IMEC, Kapeldreef 75, Leuven, B-3001, Belgium. (e-mail: Saeed.Mahmoudpour@vub.be)}
\thanks{Myllena A. Prado and Myl\`ene C. Q. Farias are with the Department of Computer Science, Texas State University, US. (e-mail: myllena@txstate.edu; mylene@ieee.org)}
\thanks{Gi-Mun Um is with the Electronics and Telecommunications Research Institute, Daejeon, Korea. (e-mail: gmum@etri.re.kr)}
\thanks{Ismael Seidel and Leonardo de Sousa Marques are with the Embedded Computing Lab., Dept. of Computer Science and Statistics, Federal University of Santa Catarina, Brazil. (e-mail: ismael.seidel@ufsc.br; leonardo.sm@grad.ufsc.br)}
\thanks{Leonardo Andrade is with the Licks Attorneys and the Federal University of Rio de Janeiro (UFRJ), Rio de Janeiro, Brazil. (e-mail:leonardo.andrade@lickslegal.com.)}
\thanks{Shengyang Zhao is with the Ningbo Institute of Digital Twin, Eastern Institute of Technology, Ningbo, China. (e-mail: szhao@idt.eitech.edu.cn)}
\thanks{Carla L Pagliari is with the Instituto Militar de Engenharia-IME, PGEE/PGED, Brazil. (e-mail: carla@ime.eb.br)}}


\markboth{Vol.~xx, No.~x, July~2026}%
{Shell \MakeLowercase{\textit{et al.}}: Bare Demo of IEEEtran.cls for IEEE Journals}



\maketitle


\begin{abstract}
Benchmarking immersive media coding solutions, especially in the standardization context, requires reliable and reproducible subjective quality assessment (QA) procedures, along with objective quality metrics that remain accurate across different distortion types. This paper presents a standardized workflow for light field QA, developed and deployed in the context of JPEG Pleno standardization activities, which integrates benchmark generation, a hybrid subjective evaluation, and objective metric analysis into a common workflow. The benchmark is designed to encompass not only traditional coding-only artifacts but also distortions that arise in processing pipelines in which light field encoding is accompanied with view synthesis and reconstruction techniques. A hybrid subjective method is proposed enabling fine-grained assessment by combining reference-anchored quality rating with targeted pairwise refinement in perceptually ambiguous regions. The reliability of subjective scores is verified using statistical consistency analyses between observers of two cohorts. Finally, a large set of objective metrics is systematically evaluated in terms of global prediction accuracy, local agreement in ambiguous quality regions, and robustness across distortion families. The results show that several metrics achieve strong agreement for coding-only stimuli, but their performance consistently drops when view synthesis distortions are included. The analysis further highlights the importance of view-pooling strategy in the design of future light field quality metrics. The work provides a reproducible and standardization-ready framework for fine-grained light field QA, while identifying key limitations of current objective metrics under emerging coding pipelines. The subjectively annotated dataset is publicly available at~\url{https://plenodb.jpeg.org/lfqa/objectivecfp}.
\end{abstract}


% \fi
\IEEEpeerreviewmaketitle

\section{Introduction}

% \textcolor{red}{Mylene please review and add references when needed. @Carla we may need 1 paragraph or less to introduce LFs}

%\textcolor{red}{Please also note this: For the initial submission of a Regular Paper, the manuscript may not exceed 13 double-column pages (using a 10-point font), including title; names of authors and their complete contact information; abstract; text; all images, figures and tables, appendices and proofs; and all references. Supplemental materials and graphical abstracts are not included in the page count. (NOTE: For the IEEE Transactions on Multimedia, the initial submission of a Regular Paper may not exceed 10 pages).}

\IEEEPARstart{L}{ight} field (LF) imaging extends conventional photography by capturing both the spatial and angular distribution of light, enabling a more complete 3D scene representation. This supports immersive free-viewpoint rendering, photorealistic view synthesis, depth estimation, and virtual or augmented reality~\cite{levoy2006light}.~LF imaging is grounded in the plenoptic function~\cite{adelson91plenoptic, Levoy:1996}, which models radiance as a function of position, direction, wavelength, and time. Because acquiring the full 7D plenoptic function is impractical, it is typically reduced to a 4D LF parameterized by $(s,t,u,v)$, where $(s,t)$ are viewpoint coordinates on one plane and $(u,v)$ are coordinates of the corresponding light ray on a parallel plane~\cite{Levoy:1996}. In practice, a 4D LF is sampled as an array of sub-aperture images encoding spatial and angular information, providing far richer visual content than conventional 2D images. The richer representation of LFs requires much larger data volumes, making efficient acquisition, compression, transmission, reconstruction, and quality assessment (QA) fundamental challenges for practical LF systems. 

Unlike conventional images, LFs exhibit redundancies both within each spatial view and across neighboring angular views, requiring coding algorithms that jointly exploit spatial and angular correlations while preserving perceptual quality. These challenges led to dedicated standards, most notably the JPEG Pleno framework~\cite{Pleno-2}. JPEG Pleno defines a standardized architecture for emerging imaging modalities, including light fields, point clouds, and holography. Its integrated design covers coding technologies and advanced functionalities such as plenoptic data manipulation, rich metadata, interactive random access, and flexible file formats. The framework also addresses visual quality evaluation, recognizing that reliable QA is critical for developing, optimizing, and benchmarking compression algorithms and other LF processing methods~\cite{mahmoudpour2026benchmarking}. 

Visual QA is of fundamental importance for LF coding, as visual degradation in LF content extends well beyond simple spatial coding artifacts (e.g. blur, blocking) within individual views. To rigorously analyze how different types of distortions influence the perceived quality of rendered LFs, a substantial body of recent work has proposed both subjective and objective QA methodologies specifically tailored to this modality~\cite{alamgeer2023survey, mahmoudpour2021performance, viola2018quality, mahmoudpour2026benchmarking}. These methods, among other aspects, evaluate angular consistency, parallax, geometric accuracy, and faithful view-dependent appearance attributes, all of which become particularly critical when sparsely sampled data are compressed and subsequently reconstructed into dense LFs.

More specifically, objective quality metrics are essential for LF coding and reconstruction pipelines, as they support codec development, parameter optimization, and large-scale performance analysis~\cite{alamgeer2023survey}. Their reliability however, depends on evaluation against perceptually valid subjective ground truth collected under representative distortion conditions. Therefore, stress testing objective metrics require diverse test stimuli~\cite{shafiee2023datasets} and subjective methods that are reliable, reproducible, and sensitive to small perceptual differences~\cite{mahmoudpour2026benchmarking}.

The need for reproducible QA has also been recognized in standardization activities for plenoptic imaging. To address this, the new Part 7 of the JPEG Pleno is focused on light field quality assessment, aiming to define common procedures for evaluating decoded LF content. Within this activity, subjective assessment was first treated as a foundational step, since objective metrics require reliable perceptual ground truth. Earlier work therefore focused on comparing subjective methodologies for LF coding assessment, including rating- and ranking-based protocols, to understand their reliability and discriminability as detailed in~\cite{mahmoudpour2026benchmarking}.

Previous research has demonstrated that conventional rating methods, such as Absolute Category Rating (ACR) and the Double Stimulus Impairment Scale (DSIS)~\cite{itur_bt500_15, itut_p910_2023}, offer valuable reference anchoring for subjective QA, while they may exhibit limited sensitivity when evaluating high-quality stimuli or conditions with small quality differences~\cite{rania2018subjective, mahmoudpour2026benchmarking}. In contrast, pairwise comparison (PC) typically yields higher discriminability, albeit at the expense of substantially increased experimental complexity and cost~\cite{mikhailiuk2021active, shah2016estimation, pastor2024comparison}. This motivated the development of a hybrid subjective methodology that combines a coarse reference-anchored rating stage with selective PC in perceptually ambiguous regions. This hybrid approach is being developed within the ongoing JPEG Pleno Part 7 standardization work on subjective light field quality assessment. Building on this subjective QA foundation, the activity is subsequently being complemented by a call for proposals (CfP) on objective quality metrics, supporting the development of standardized objective LF QA tools.

The present paper builds on this subjective QA phase~\cite{mahmoudpour2026benchmarking} and addresses the next step: the implementation of a framework for fine-grained evaluation of the reliability of objective quality metrics for LFs. The proposed workflow includes source-scene selection, stimulus generation, subjective QA using the hybrid protocol, score processing procedure, and metric analysis. The framework serves as a reproducible baseline for analyzing metric behavior across distortion families and scene characteristics, laying the groundwork for a CfP on objective metrics. The aim is therefore not to report or compare submitted metric proposals in this paper, but to present and analyze the standardized test procedure itself as a methodological basis for the development and assessment of objective metrics for LFs.

%A central challenge in designing such a procedure is that the evaluation material must be sufficiently broad to represent the types of degradations that objective metrics are expected to predict. Existing LF quality datasets and evaluation studies often focus on isolated distortion families, such as compression, angular subsampling, or view synthesis, which is valuable for controlled analysis but does not fully reflect emerging end-to-end processing pipelines. In practice, LF content may be sparsely acquired or transmitted, compressed at reduced angular density, and then expanded again through interpolation or neural reconstruction. The resulting quality degradation is therefore not caused by a single operation, but by the interaction between compression artifacts and reconstruction or interpolation errors. To account for this, the test material considered in this work includes conventional coding-only conditions as well as sparse-to-dense conditions in which decoded sparse LFs are reconstructed using interpolation and 3D Gaussian Splatting (3DGS). This design enables the evaluation procedure to probe whether objective metrics remain reliable when artifacts are spatially localized, angularly inconsistent, or dependent on view synthesis.

The remainder of the paper is organized as follows. Section~\ref{sec:background_motivation} reviews subjective and objective LF quality assessment methods and discusses current challenges in metric evaluation. Section~\ref{sec:dataset_stimuli} describes the source scenes, LF representations, and stimulus-generation procedure. Section~\ref{sec:subjective_procedure} presents the hybrid subjective test protocol, including test setup, observer screening, and score reconstruction. Section~\ref{sec:metric_eval_protocol} defines the objective metric evaluation protocol, including anchor metrics, score aggregation, and statistical evaluation criteria. Section~\ref{sec:benchmark_results} reports the subjective and objective benchmark results. Section~\ref{sec:disc} discusses the implications for objective LF metric design, and Section~\ref{sec:conc} concludes the paper.

%The main contribution of this work is the development and deployment of a standardized evaluation procedure for objective LF quality metrics. The procedure covers the complete chain from source material selection to metric analysis: a diverse set of LF scenes is selected, test stimuli are generated using coding-only, interpolation-based, and 3DGS-based processing conditions, subjective ground truth is collected using a proposed hybrid protocol, perceptual scores are processed through a consistent statistical model, and objective metrics are evaluated against the resulting scores. This end-to-end design ensures that metric performance is assessed under controlled yet diverse conditions, including distortions that go beyond conventional compression artifacts.

\section{Background and Motivation}
\label{sec:background_motivation}
% \textcolor{red}{@Mylene:could you please take care of the QA background in this section? I tried to already add some text but feel free to change. we already had an extensive background in the previous journal paper so here i would not repeat but keep shorter, more towards standardization, and refer to our paper when necessary}

%\textcolor{red}{@Carla: (a) Maybe a short intro first to the plenoptic modalities, light fields and standardization HERE? (b) LF should also briefly introduced in the Introduction}

% Carla 

% LFs 7D to 4D

% ---------------------------------------------------------------
\subsection{Subjective Quality Assessment Methods}
\label{sec:extended_visual_stimuli}

Visual quality can be evaluated by subjective QA, where humans rate media quality under controlled conditions, or by objective QA, which computationally predicts human judgments. Subjective QA typically follows standardized protocols~\cite{itur_bt500_15, itut_p910_2023} that maximize measurement precision, consistency, and reliability. Developed mainly for images and videos, these standards define the environment and apparatus, test procedures, and criteria for selecting and preparing test content. Methods are generally classified as rating-based or ranking-based (comparative judgment).

Within the class of rating-based assessment protocols, the Double Stimulus Continuous Quality Scale (DSCQS) and the Double Stimulus Impairment Scale (DSIS) are among the most frequently employed. In these methods, test and reference stimuli are presented as paired sequences, and observers provide quality judgments for the test stimulus relative to its corresponding reference. Ranking-based protocols, such as pairwise comparison (PC)~\cite{david1963method}, typically provide higher discriminative power and better statistical robustness~\cite{perez2019pairwise} than rating-based procedures~\cite{shah2016estimation}. PC produces an ordering of stimuli by repeatedly presenting pairs and counting how often one is preferred. However, the resulting perceptual distances are purely relative and do not capture fidelity to an absolute reference~\cite{pastor2023discriminability, pastor2024comparison}. Thus, the derived scales, though often highly predictive of perceived quality~\cite{watson2001measurement}, lack direct semantic meaning in terms of absolute quality levels. A common solution is to use triplet-based schemes, where two test stimuli are evaluated together with a reference~\cite{pastor2024comparison, men2021subjective}. PC-based studies are also time-consuming due to the large number of comparisons, motivating approaches such as active sampling to reduce and streamline experiments~\cite{ling2020strategy, mikhailiuk2021active}.

Light field (LF) QA is a demanding and complex subarea of multimedia QA. While LFs can be modeled as collections of sub-aperture images, their perceived quality depends on more than the spatial fidelity of individual views. Angular consistency, parallax stability and view-dependent appearance all shape perceptual quality during interactive viewpoint navigation and become particularly critical when LF content undergoes compression, angular subsampling, interpolation, or other reconstruction processes.

A substantial body of research has investigated subjective visual QA of LFs~\cite{mahmoudpour2026benchmarking, alamgeer2023survey, viola2017new, viola2018quality}, deepening understanding of the complex challenges in evaluating immersive LF media. A key difficulty is designing a stimulus presentation protocol that faithfully reproduces LF exploration from multiple viewpoints. Most studies use pseudo-video sequences or interactive multiview presentations to convey angular parallax and depth cues, typically combined with rating-based evaluation schemes.

Although dynamic LF visualizations (e.g. pseudo-video or interactive rendering), are effective for rating-based methods with a fixed reference, extending them to a triplet-based paradigm for reference-anchored PCs greatly increases observers’ cognitive load~\cite{saraiva2025subjective, mahmoudpour2026benchmarking}. Because LFs require much longer viewing times than 2D images, exhaustive PC experiments become prohibitively time-consuming. This limitation is even more pronounced in dynamic triplet-comparison frameworks, which further increase task complexity and often make the procedure impractical or uncomfortable for subjective QA.

The complementary strengths of rating- and ranking-based methods motivate a hybrid protocol. Ratings provide reference-based judgments on interpretable scales, while ranking methods, including PC, offer higher sensitivity and discrimination. Here we propose a systematic hybrid framework that leverages the strengths of both. In this design, first, a coarse Double Stimulus Comparison Scale (DSCS) defines the global quality structure of stimuli relative to the reference. Next, PC is applied only to stimuli within the same quality category, where ratings lack resolution. This preserves the semantic clarity of ratings while recovering fine-grained perceptual order in ambiguous regions. Finally, psychometric reconstruction fuses both information sources into a continuous quality scale using Thurstone scaling and a unified analysis of PC and rating data~\cite{perez2019pairwise}. A detailed analytical assessment of the hybrid protocol is provided in~\cite{Saeed_2026_LFQA}, while the present paper reports its practical implementation and suitability for fine-grained objective metric assessment.

\subsection{Objective LF Quality Metrics}
\label{sec:subjective_to_objective}
While objective QA is well studied for conventional videos and images~\cite{min2024perceptual, shahid2014no, zheng2024video}, assessing immersive media quality, especially LF images, remains a distinct and largely unsolved problem. Conventional 2D QA methods ignore the unique, multidimensional characteristics of such content~\cite{Shi2018PerceptualLF, amirpour2019reliability}. Unlike conventional image quality metrics, LF metrics must consider both the quality of individual sub-aperture images and the coherence of angular information across viewpoints. Compression- or reconstruction-induced distortions jointly affect spatial detail, depth perception, inter-view consistency, and motion parallax. Thus, traditional 2D image QA metrics such as PSNR and SSIM are poor predictors of perceived quality in LF imagery. 

In this work, we focus on full-reference (FR) methods, which are particularly suitable for compression evaluation, where the objective is to preserve the highest possible fidelity with respect to the reference content. Early LF QA research mainly tested conventional image quality metrics on LF data. Adhikarla \emph{et al.} thoroughly evaluated image, video, and multiview metrics on dense LFs, finding that while traditional metrics perform reasonably when an undistorted reference is available, they fail to reliably capture angular artifacts and inter-view inconsistencies between neighboring perspectives~\cite{Adhikarla_2017_CVPR}.

Subsequent work introduced handcrafted FR metrics tailored to LFs, typically combining spatial, angular, and geometric information. Tian \emph{et al.}~\cite{ Tian2020SDFM, Tian2020LFC} proposed the Symmetry and Depth Feature-based Model (SDFM), and the LF Coherence (LGF-LFC) metric, which use derivative features, symmetry descriptors, depth cues, and epipolar plane image (EPI) coherence to better capture perceptual distortions. Meng \emph{et al.}~\cite{mengfrlfiqa2020} further leveraged angular-spatial characteristics via focus stacks, while Huang \emph{et al.}~\cite{Huang2021CTM} introduced contourlet transform- and spatial-geometry–based models to jointly analyze texture and geometric consistency. By explicitly modeling LF–specific properties, these methods consistently outperform classical 2D metrics, underscoring the need to preserve both spatial fidelity and angular coherence.

Another research direction integrates multiple perceptual cues into unified quality models. Min \emph{et al.}~\cite{Min2020LFMetric} proposed a metric that combines global spatial quality, local structural degradation, and angular consistency to assess distortions from LF reconstruction, compression, and display. Similarly, Ma \emph{et al.}~\cite{Ma2023NSS} used macro-pixel representations with natural scene statistics and texture degradation features to separate spatial and angular information before regression-based quality prediction. These methods acknowledge that no single feature can represent the diverse distortions in practical LF processing pipelines.

Deep learning has significantly advanced LF-IQA by enabling feature representations to be learned directly from data rather than relying on manually engineered descriptors. Zhang \emph{et al.}~\cite{Zhang2023EDDMF} proposed EDDMF, a convolutional neural network that learns hierarchical discrepancy features between reference and distorted LF patches, achieving improved prediction accuracy while maintaining moderate computational complexity. Transformer-based architectures have further enhanced performance by modeling long-range dependencies across both spatial and angular dimensions. The multidimensional attention network proposed by Zhang \emph{et al.}~\cite{Zhang2025Attention} leverages attention mechanisms to jointly capture local texture degradation and global view correlations.

\subsection{Challenges of LF Quality Assessment}
\label{sec:lfqa_challenges}

Despite progress in LF QA, several challenges remain for reliable assessment. Many existing subjectively-annotated LF datasets are limited in spatio-angular characteristics, or distortion diversity. In particular, several benchmarks are built around narrow LF acquisition configurations, such as lenslet content with limited angular baselines, which may not sufficiently stress-test metrics under broader and more diverse LF conditions. Other datasets focus on isolated distortion types, such as compression, angular subsampling, or view synthesis. While these controlled settings are useful, they do not fully reflect modern LF processing pipelines, where content may be sparsely captured or transmitted, compressed at reduced angular density, and then reconstructed into dense angular representations. In this case, perceived quality may be affected by the interaction of compression, interpolation, and neural reconstruction artifacts rather than by a single degradation source. Therefore, objective metrics should be tested on datasets that include both conventional coding artifacts and sparse-to-dense reconstruction distortions, including interpolation and recent 3D Gaussian Splatting (3DGS)-based processing~\cite{kerbl20233d}.

A further challenge concerns the practical usability and generalizability of existing LF-specific metrics. Public implementations are often unavailable, limiting reproducibility and making systematic cross-dataset benchmarking difficult. Our previous work~\cite{mahmoudpour2026benchmarking, mahmoudpour2021performance} summarizes the characteristics of existing subjective LF datasets and highlights the limited generalizability of current LF-specific metric families. Consequently, despite the limitations of conventional 2D image and video quality metrics for LF content, they remain the most practical and reproducible baselines for large-scale evaluation. 

In this work, an evaluation framework is designed to provide reliable perceptual ground truth for fine-grained objective metric assessment under modern LF compression pipelines. The paper also reports the performance of a set of anchor objective metrics on the proposed dataset, providing a reproducible baseline for future comparison. Given the observed lack of a well-established metric for LFs, the standardization efforts presented in this paper, ultimately aim to establish an assessment pipeline, together with LF quality metrics, that is reliable and generalizable across diverse LF processing conditions, openly accessible to the research community, and suitable for consistent evaluation in practical coding workflows.

\section{Evaluation Dataset and Stimulus Generation}
\label{sec:dataset_stimuli}

The evaluation dataset was designed to assess objective quality metrics across diverse LF degradations. It includes both coding-only stimuli and densely sampled LFs reconstructed from sparsely coded views. This allows the benchmark to test metric reliability not only for compression artifacts but also for distortions introduced by interpolation and neural reconstruction pipelines.

\subsection{Source Light Field Scenes}
\label{sec:source_scenes}
\begin{figure*}[t]
\centering
\begin{subfigure}[t]{0.64\textwidth}
    \centering
    \includegraphics[width=\linewidth]{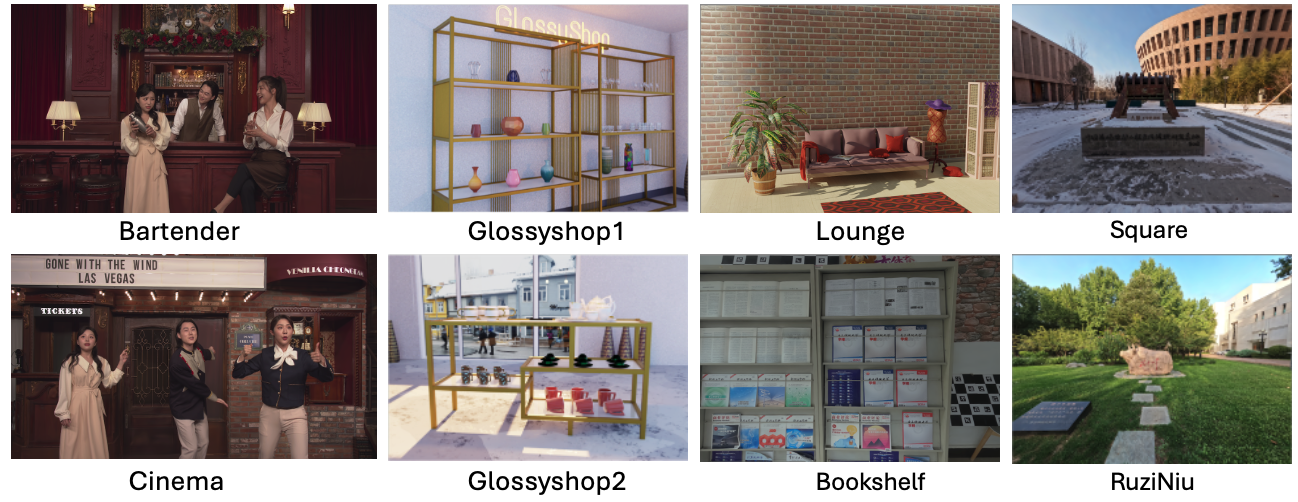}
    \caption{Representative views of the eight source light field scenes.}
    \label{fig:source_lf_views_examples}
\end{subfigure}
\hfill
\begin{subfigure}[t]{0.33\textwidth}
    \centering
    \includegraphics[width=\linewidth]{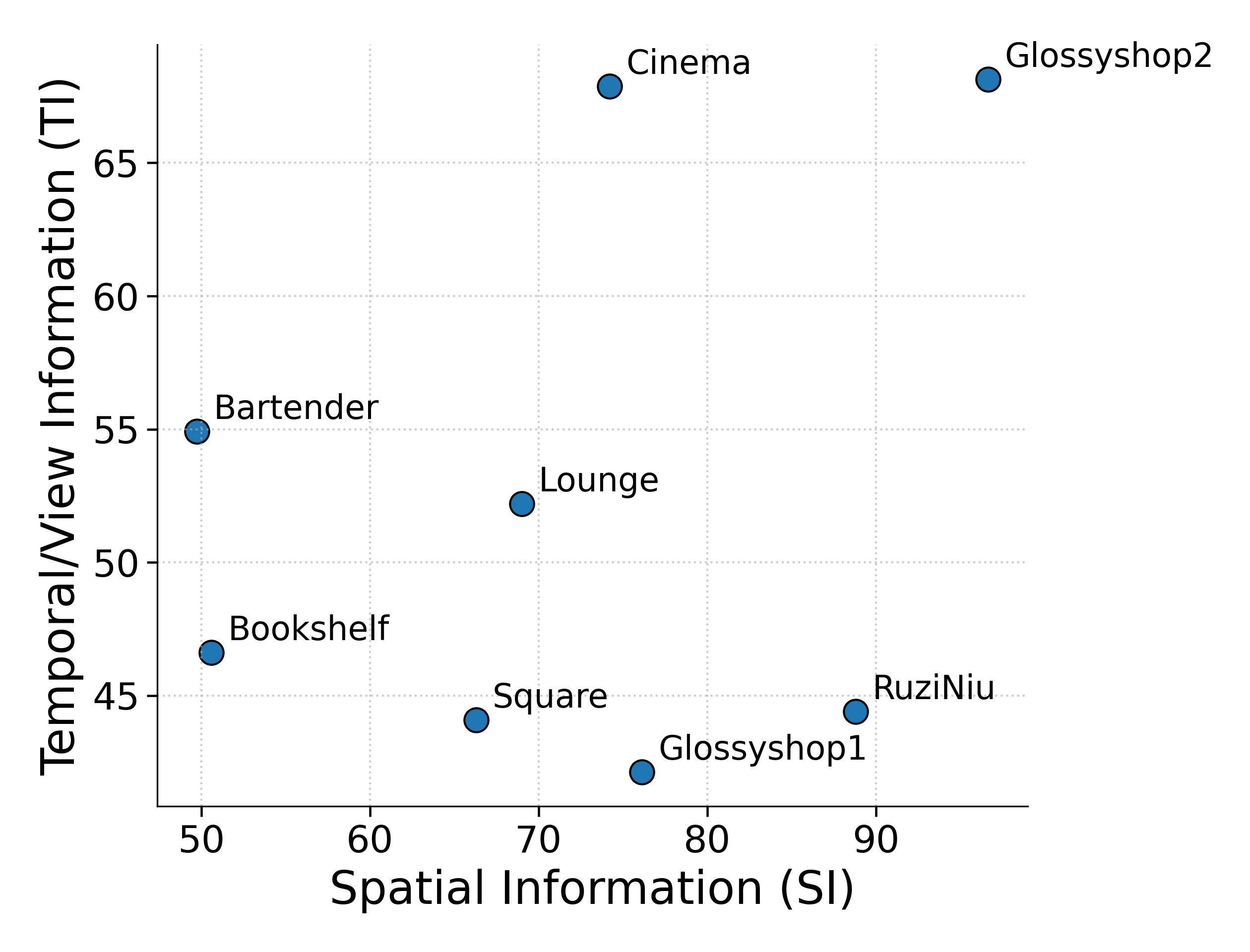}
    \caption{SI--TI distribution of the scenes.}
    \label{fig:source_lf_views_siti}
\end{subfigure}
\caption{Source LF scenes used in the benchmark. (a) Representative views. (b) Spatial information (SI) and temporal/view information (TI) characterize the diversity of contents.}
\label{fig:source_lf_views}
\end{figure*}

The source material was selected to cover a broad range of LF characteristics, including spatial resolution, angular sampling density, baseline, scene geometry, texture complexity, and non-Lambertian surfaces. Eight source scenes were included in the final evaluation dataset, comprising natural indoor and outdoor captures and synthetic scenes. Representative views of the selected scenes are shown in Fig.~\ref{fig:source_lf_views}(a), and their spatial and angular resolutions are summarized in Table~\ref{tab:source_scenes}.

The \textit{Bartender} and \textit{Cinema} scenes are natural indoor content captured with multi-camera arrays with converging optical axes. They feature human-centric content and moderate-to-large baselines, posing challenges for both compression and view synthesis. A 2D camera array with horizontal and vertical baselines of about 50~cm and 25~cm was used. %The scenes were originally acquired on sparse grids and later resampled to denser angular representations via view synthesis.

The scenes \textit{RuziNiu}, \textit{Square}, and \textit{Bookshelf} were acquired using a compound-eye multi-camera capture system. These captures provide dense image collections that were subsequently rendered into regular LF grids. They complement the multi-camera array captures by providing different spatial resolutions and angular sampling patterns. 

The synthetic scenes \textit{Glossyshop1}, \textit{Glossyshop2}, and \textit{Lounge} were rendered in Blender to complement real-world data with controlled variations in material properties and lighting. The \textit{Glossyshop} scenes emphasize specular reflections and non-Lambertian effects, while \textit{Lounge} features smoother surfaces and simpler structures. The inter-camera spacing was 3~cm, with virtual cameras using a 35.0~mm focal length and 32.0~mm sensor width. Depth of field was set with an aperture of $f/100.0$ and an 8.0~m focus distance.

To assess the diversity of the selected source scenes, Fig.~\ref{fig:source_lf_views}(b) shows their spatial information (SI) and temporal/view information (TI). SI measures spatial detail and edge activity in the reference views, while TI captures view-to-view variation across the ordered LF views. Because the content is static, TI reflects angular/viewpoint changes due to parallax, occlusions, and scene geometry rather than physical motion. The SI--TI distribution shows that the benchmark spans a wide range of texture complexity and angular variation, which is crucial for objective metric evaluation, as metric behavior can strongly depend on both spatial detail and inter-view changes.

%\begin{figure}[t]
%\centering
% \includegraphics[width=\linewidth]{figures/dense_sparse_grid.pdf}
%\caption{(a) Illustration of dense and sparse angular representations. (b) Representative examples of different distortion families included in the benchmark: coding-only degradation, interpolation-based reconstruction artifacts, and 3DGS-based reconstruction artifacts.}
%\label{fig:dense_sparse_grid}
%\end{figure}

\begin{table}[t]
\centering
\caption{Source LF scenes and final angular configurations used for benchmark generation.}
\label{tab:source_scenes}
\footnotesize
\setlength{\tabcolsep}{6pt}
\renewcommand{\arraystretch}{1.15}
\begin{tabular}{lcccl}
\toprule
\textbf{Scene} 
& \makecell{\textbf{Spatial}\\\textbf{resolution}} 
& \makecell{\textbf{Dense}\\\textbf{grid}} 
& \makecell{\textbf{Sparse}\\\textbf{grid}} 
& \makecell{\textbf{Content}\\\textbf{type}} \\
\midrule
Bartender   & $1920 \times 1080$ & $25 \times 4$  & $7 \times 4$ & Natural indoor \\
Cinema      & $1920 \times 1080$ & $25 \times 4$  & $7 \times 4$ & Natural indoor \\
RuziNiu     & $1600 \times 1190$ & $15 \times 15$ & $8 \times 8$ & Natural outdoor \\
Square      & $1600 \times 1200$ & $15 \times 15$ & $8 \times 8$ & Natural outdoor \\
Bookshelf   & $1600 \times 1200$ & $15 \times 6$  & $8 \times 6$ & Natural indoor \\
Glossyshop1 & $1082 \times 750$  & $25 \times 13$ & $7 \times 7$ & Synthetic \\
Glossyshop2 & $1082 \times 750$  & $25 \times 5$  & $7 \times 3$ & Synthetic \\
Lounge      & $1082 \times 750$  & $25 \times 13$ & $7 \times 7$ & Synthetic \\
\bottomrule
\end{tabular}
\end{table}

\subsection{Dense and Sparse Light Field Representations}
\label{sec:dense_sparse_lfs}

For each source scene, we used dense and sparse angular representations. The dense representation defines the reference LF sampling grid for evaluation and visualization, while the sparse one is a reduced set of viewpoints used in additional coding and reconstruction experiments. This distinction is central to the benchmark, enabling analysis of (i) distortions from direct LF coding and (ii) distortions from reconstructing densely sampled LFs from compressed sparse inputs. It also reflects realistic acquisition and transmission constraints on angular sampling and provides controlled conditions to assess how objective quality metrics respond when compression artifacts combine with reconstruction or interpolation errors.

Sparse acquisition configurations were generated via structured angular subsampling. For \textit{Bartender}, \textit{Cinema}, and \textit{Bookshelf}, subsampling was applied along the horizontal angular axis, keeping fewer angular columns while preserving full vertical resolution. In contrast, \textit{RuziNiu}, \textit{Square}, and the remaining synthetic scenes were subsampled along both angular dimensions. The resulting sparse angular grids preserve enough directional information for accurate reconstruction while reducing the number of views to transmit or encode.

\subsection{Coding-only Stimuli}
\label{sec:coding_only_stimuli}
%\textcolor{red}{Ismael and Carla. I made a first draft, feel free to revise}

Coding-only stimuli were created to serve as a conventional compression baseline for the benchmark. Two coding strategies were employed: JPEG PLeno Model (JPLM)~\cite{JPLM}, using its 4D Transform Mode (4D-TM), and x265/HEVC~\cite{x265}. A 4D LF can be represented as a 2D grid of images (views). The ($s\times t$) coordinates specify the individual 2D views, while the pixel positions within each view are given by the ($u\times v$) coordinates~\cite{Levoy:1996}. The x265/HEVC encoder applies pseudo-video coding to the $s\times t$ views in an IPPP serpentine scan order (left-to-right, top-to-bottom), as specified in the JPEG Pleno Common Test Conditions (CTC)~\cite{CTC_JPEG_Pleno_v34}. In this scheme, the x265/HEVC pipeline rearranges the 4D LF data into a sequence of 2D frames so that it can be handled by conventional codecs originally developed for 2D video content. In contrast, the 4D-TM codec is a 4D-native LF coding method that jointly exploits the full redundancy of the LF across all four dimensions, i.e., both between and within views. Consequently, the two codecs leverage 4D redundancy in fundamentally different manners, resulting in distinct types of coding artifacts.

All coding/decoding procedures used the JPEG Pleno Light Field Coding Toolkit~\cite{pleno_LFC_toolkit}. This modular toolkit is designed to provide functionalities such as file format conversion, encoding, and decoding with selected codecs, and quality assessment, ensuring all tasks are performed within a unified pipeline to avoid inconsistencies in results. It can be used to compare its outputs with those of other methods for benchmarking and evaluation purposes. The coding-only track uses five bitrate points per codec and scene, covering low to high quality with fine steps to capture slight perceptual differences. In total, 80 coding-only stimuli were generated for the final benchmark.

\subsection{Coding and Reconstruction from Sparse Light Fields}
\label{sec:sparse_reconstruction}

In practical LF coding, the full angular content is often not coded directly. Instead, a sparse subset of views is encoded and transmitted, and a dense angular representation is reconstructed at the decoder. This is important for the proposed benchmark because final perceived quality depends not only on compression artifacts but also on how they interact with interpolation or reconstruction errors. We considered two dense reconstruction strategies: view interpolation and 3DGS-based reconstruction. Both operate on decoded sparse LFs to produce dense angular representations aligned with the target grids, enabling analysis of metric behavior under spatially localized, angularly inconsistent, or synthesis-dependent distortions.

\subsubsection{Interpolation-based Reconstruction}
\label{sec:interpolation_dense_reconstruction}

Missing views between decoded sparse samples were synthesized using interpolation. Two methods were used, RIFE~\cite{huang2022real} and SepConv++~\cite{Niklaus_WACV_2021}, with the choice per scene based on visual suitability and baseline. SepConv++ was used for narrow-baseline scenes where preserving local structure is critical, while RIFE was used where motion/flow-based interpolation was more stable. For JPLM, two bitrate points per scene were kept; for x265, three bitrate points were selected, including a high-bitrate anchor so that artifacts are dominated by interpolation. This interpolation track yields 40 stimuli in the final benchmark.

\subsubsection{3DGS Reconstruction}
\label{sec:3dgs_dense_reconstruction}

Modern reconstruction was implemented with 3DGS to introduce artifacts distinct from conventional interpolation and compression. The pipeline estimates camera parameters, trains a 3DGS model from decoded sparse views, and renders the target dense LF grid. Sparse LFs are first decoded at selected bitrates using JPLM or x265, and missing views are then reconstructed with 3DGS. This track was generated for six scenes with stable camera pose estimation and reconstruction: \textit{Bartender}, \textit{Cinema}, \textit{Bookshelf}, \textit{Glossyshop1}, \textit{Glossyshop2}, and \textit{Lounge}. Two bitrate points were kept per codec and per scene, yielding 24 stimuli. The reconstruction and interpolation tracks add 64 stimuli, which, with the 80 coding-only stimuli, yield 144 stimuli in total. %This design enables a comprehensive analysis of distortions arising from both conventional compression and modern reconstruction pipelines.

%\subsection{Final Benchmark Composition}
%\label{sec:benchmark_composition}
%\textcolor{red}{Saeed}
%The final benchmark combines coding-only, interpolation-based, and 3DGS-based stimuli. This composition was selected to provide a controlled but diverse set of distortion conditions for objective metric evaluation. Coding-only stimuli provide a conventional compression baseline, interpolation-based stimuli introduce sparse-to-dense view synthesis artifacts, and 3DGS-based stimuli introduce neural reconstruction artifacts. Together, these categories allow anchor metrics to be evaluated not only in terms of global correlation with subjective scores, but also in terms of robustness across distortion families.

%This final set of 144 stimuli serves as the basis for the hybrid subjective assessment and subsequent objective metric analysis.

\section{Hybrid Subjective Evaluation Procedure}
\label{sec:subjective_procedure}
% \textcolor{red}{Saeed will draft, Mylene to review}

%\subsection{Subjective Test Design}
%\label{sec:subjective_test_design}

\begin{figure*}[t]
\centering
\includegraphics[width=0.85\textwidth]{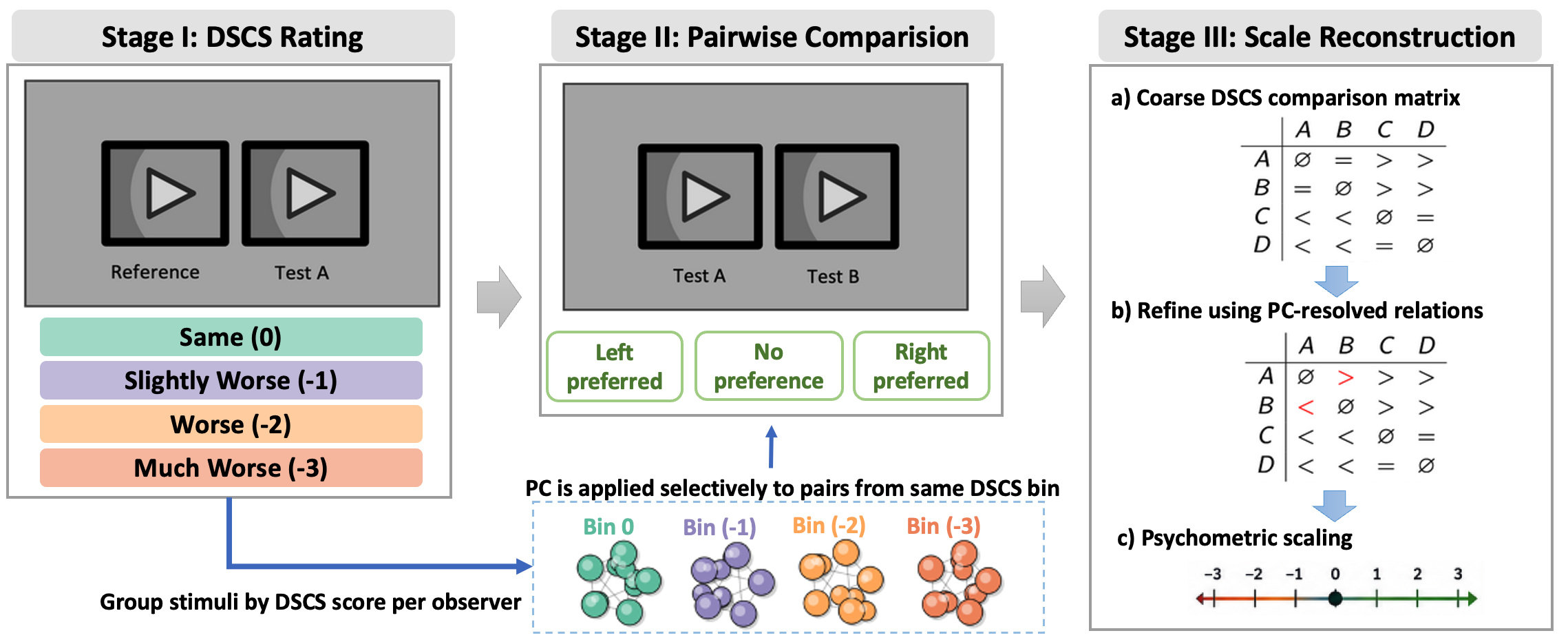}
\caption{Overview of the hybrid DSCS+PC subjective assessment protocol. In Stage~I, each distorted stimulus is rated relative to the reference on a four-level DSCS scale, and stimuli are grouped by observer-specific DSCS scores. In Stage~II, pairwise comparisons are selectively applied within each DSCS bin to resolve local ambiguities. In Stage~III, the DSCS comparison matrix is refined using these PC results and psychometrically scaled into continuous perceptual quality scores.}
\label{fig:hybrid_dscs_pc_framework}
\end{figure*}

The subjective assessment protocol used in this study follows the hybrid methodology developed in the \textit{JPEG Pleno Part 7: Light Field Quality Assessment} standardization activities. This approach combines the Double Stimulus Comparison Scale (DSCS) with Pairwise Comparison (PC), hereafter DSCS+PC. DSCS is a reference-based, double-stimulus procedure in which each impaired stimulus is rated relative to its reference, whereas PC presents two impaired stimuli and asks observers to choose the preferred one.

Fidelity-oriented LF-QA needs subjective scores that are both interpretable and discriminative. Rating-based methods express perceived degradation relative to a reference but lose sensitivity when several stimuli fall into the same or neighboring quality categories. This is especially limiting in benchmarks that include fine-grained coding levels. In such cases, PC generally offers better perceptual sensitivity and finer discriminative power between quality levels.

Our hybrid design tackles a key challenge in subjectively evaluating extended visual stimuli. In dynamic, immersive, or multi-view assessment, observers explore content over time or across viewpoints rather than judging single still images. Exhaustive PC is highly discriminative but does not scale, while rating-based methods are practical but less sensitive, especially in ambiguous quality ranges. The hybrid DSCS+PC approach resolves this by applying PC only where extra discrimination is most valuable, while preserving the semantic clarity and reference anchoring of DSCS ratings. Although demonstrated for LF QA, the framework is general and applies to videos and other extended visual stimuli where full pairwise test is impractical.

Fig.~\ref{fig:hybrid_dscs_pc_framework} overviews the hybrid workflow, which includes two subjective stages followed by scale reconstruction. In Stage~I, observers conduct a DSCS assessment, rating each distorted stimulus relative to its reference on a four-level ordinal scale: \emph{Same} ($0$), \emph{Slightly worse} ($-1$), \emph{Worse} ($-2$), and \emph{Much worse} ($-3$). This yields a coarse but semantically meaningful degradation scale and defines observer-specific groups of stimuli mapped to the same perceived quality category.

In Stage~II, PC is used selectively to resolve locally uncertain regions. For each observer, PC trials include only stimulus pairs that received identical DSCS scores from that observer in Stage~I. Thus, the PC stage is locally focused and observer-specific, used only when that observer’s initial ratings did not distinguish between stimuli.

In Stage~III, DSCS and PC responses are combined into a continuous perceptual degradation scale. DSCS provides the global reference-based structure, while selective PC responses refine the ordering of stimuli in ambiguous same-category regions. The resulting DSCS+PC scores preserve the interpretability of reference-based ratings, increase sensitivity to subtle perceptual differences, and remain practical for large LF benchmarks. To further limit the test duration into one 20-25 min session, the selective PC stage was restricted to pairs from the \emph{Slightly worse} ($-1$) and \emph{Worse} ($-2$) categories, which typically contain medium- to high-quality stimuli with subtler perceptual differences and a higher risk of rating-scale saturation, as also found in our previous study~\cite{mahmoudpour2026benchmarking}. The number of PC trials was capped at 200 comparisons per observer, retaining on average 82\% of eligible comparisons while avoiding excessively long sessions.

\subsection{Test Setup and Participants}
\label{sec:lab_setup_participants}

Subjective experiments followed a common evaluation protocol at a primary site (Lab1), with additional cohorts from another site (Lab2). All cohorts used the same test design, stimulus presentation, response scales, training, and score processing. Final scores were obtained by pooling all valid observers after identical screening and processing. After outlier removal, the final integrated dataset included 50 observers. Lab1 had 20 observers (13 male, 7 female; mean age 30.1 years), and Lab2 had 30 observers (12 male, 18 female; mean age 29.21 years).

Naïve observers first completed visual screening. Visual acuity was verified with a Snellen chart, and normal color vision with the Ishihara test. Before the main experiment, participants completed a training session to familiarize themselves with the visualization mode, response interface, and rating and comparison tasks. Training stimuli, distinct from those in the main evaluation, were chosen to illustrate different quality levels and distortion characteristics.

All stimuli were shown on calibrated 4K 2D displays in a side‑by‑side format with 1:1 pixel mapping. Calibration used the monitors’ integrated sensors to target sRGB, a D65 white point, 140~$cd/m^2$ peak luminance, and a black level of $\leq$ 0.2~$cd/m^2$. LF views were rendered as passive video sequences in serpentine order. Viewing distance and ambient illumination followed ITU-R BT.500. The experiment consisted of two 25-minute sessions corresponding to the DSCS and selective PC stages. A dedicated interface automatically ran the hybrid procedure, generating observer-specific PC trials from each subject's DSCS responses in the first session.

\subsection{Subject Screening}
\label{sec:subject_screening}

We used a two-stage screening to remove outliers. Observer reliability was first assessed for the DSCS stage; only subjects who passed this step were included in the subsequent PC-stage reliability analysis. DSCS reliability followed ITU-R BT.500-15 \cite{itur_bt500_15} with a leave-one-subject-out procedure: for each subject, DMOS values were recomputed without that subject, and agreement between that subject’s DSCS ratings and the leave-one-subject-out DMOS was measured using $\rho_s$, the Spearman Rank-Order Correlation Coefficient (SRCC). Subjects with $\rho_s \geq 0.7$ were retained; all others were excluded from further analyses, including the PC stage.

For the PC stage, observer reliability was assessed using the likelihood-based outlier analysis in~\cite{perez2017practical}. For each observer, a consensus Thurstone Case~V perceptual scale was estimated from the pairwise responses of all other observers, and the excluded observer's PC responses were evaluated against this scale. The probability of preferring one stimulus over another depends on the difference between their latent quality scores, so responses consistent with the consensus ordering have higher likelihoods. For each observer, the mean log-likelihood of the PC responses was computed and standardized across observers; observers with standardized log-likelihoods below $z<-3$ were deemed inconsistent in the PC stage.

\subsection{Score Processing and Perceptual Scale Reconstruction}
\label{sec:score_processing}

After subject screening, DSCS and PC responses were combined to estimate final quality scores. DSCS ratings define the global, reference-anchored quality structure, and PC responses refine only pairs that were ambiguous in the DSCS stage. Final scores are obtained by mapping both sources to a common comparison space and estimating a latent perceptual scale.

For each scene $c$, let $r_{i,s}^{(c)} \in \{0,-1,-2,-3\}$ be subject $s$’s DSCS rating for stimulus $i$. A scene-wise comparison matrix is then built from these ratings. For any two stimuli $i$ and $j$, if $r_{i,s}^{(c)} > r_{j,s}^{(c)}$, $i$ is preferred; if $r_{i,s}^{(c)} < r_{j,s}^{(c)}$, $j$ is preferred. Equal ratings are treated as ties, assigning half a vote to each direction. This converts ordinal DSCS ratings into pairwise preferences while preserving the quality ordering.

An explicit reference node is added to the comparison matrix to anchor the perceptual scale. When direct PC data exist for a stimulus pair, the DSCS-implied tie is replaced by the observed response. Thus, the PC stage preserves the global ordering across DSCS categories and only refines the local ordering of stimuli within the same coarse quality label.

Let $C_{i,j}^{(c)}$ be the aggregated count of how often stimulus $i$ is preferred to $j$ in scene $c$, combining DSCS-derived relations and available PC responses. The latent quality scores $q_i^{(c)}$ are estimated with a Thurstone Case~V model, where the probability that $i$ is preferred to $j$ is
\begin{equation}
P(i \succ j) =
\Phi\left(q_i^{(c)} - q_j^{(c)}\right),
\end{equation}
and $\Phi(\cdot)$ is the standard normal cumulative distribution function. For each scene, the latent scores are estimated by maximizing
\begin{equation}
\mathcal{L}^{(c)}
=
\sum_{i<j}
C_{i,j}^{(c)}
\log
\Phi\left(q_i^{(c)} - q_j^{(c)}\right)
+
C_{j,i}^{(c)}
\log
\Phi\left(q_j^{(c)} - q_i^{(c)}\right),
\end{equation}
subject to $q_{\mathrm{ref}}^{(c)} = 0$.
The resulting scores define a continuous perceptual scale per scene. Anchored by the reference and constrained by the DSCS ordering, they retain the meaning of degradation relative to the original LFs, while the selective PC responses refine resolution in ambiguous quality regions.

\section{Objective Metric Evaluation Protocol}
\label{sec:metric_eval_protocol}

% \textcolor{red}{Saeed will draft, Mylene to review}

\subsection{Anchor Metrics}
\label{sec:anchor_metrics}

The benchmark framework systematically evaluates objective quality metrics and their behavior under coding-only and view-synthesis distortions. It focuses on FR metrics, as the benchmark targets fidelity-oriented assessment with an original reference LF for each stimulus, making FR metrics the natural choice for measuring degradation relative to the reference. Although our previous work~\cite{mahmoudpour2026benchmarking, mahmoudpour2021performance} also examined no-reference LF quality metrics, they are less suitable as primary anchors here, because many are trained or validated under narrower assumptions (e.g., lenslet content) and may not generalize reliably across datasets and distortion types.

The selected metric set includes classical FR image quality metrics (PSNR, SSIM~\cite{ssim}, MS-SSIM~\cite{MSSSIM2003}, IW-SSIM~\cite{iwssim}, VIF~\cite{vif2006}, FSIM~\cite{fsim}, GMSD~\cite{gmsd}, MAD~\cite{mad}, NLPD~\cite{laparra2017perceptually}) and deep feature-based perceptual metrics (LPIPS~\cite{lpips}, DISTS~\cite{dists}, ST-LPIPS~\cite{ghildyal2022stlpips}, DeepDC~\cite{zhu2023DeepDC}) to assess their ability to capture perceptual degradations in LF content. Video quality metrics (VMAF~\cite{netflix2016vmaf}, CVVDP~\cite{mantiuk2024colorvideovdp}) are evaluated because the LF stimuli are rendered as passive view sequences, where angular variations and view-to-view inconsistencies may matter. Finally, immersive-video extensions (IV-PSNR, IV-SSIM) are included as metrics that explicitly account for multi-view assessment.

\subsection{Metric Score Aggregation}
\label{sec:metric_score_aggregation}

Subjective scores are assigned to complete LF stimuli, so view-level metric predictions should be combined into a single stimulus-level score. The default aggregation is the arithmetic mean over views. Since interpolation and 3DGS artifacts can be confined to specific views or angular regions, we consider alternative pooling strategies. Let $m_v$ be the quality-oriented objective score (higher is better) for view $v$, $v=1,\ldots,N$. Worst-$X\%$ pooling averages the lowest-quality views:
\begin{equation}
M_{\mathrm{worst}\text{-}X}
=
\frac{1}{K}
\sum_{v \in \mathcal{W}_X} m_v ,
\end{equation}
where $\mathcal{W}_X$ contains the $K=\lceil NX/100\rceil$ lowest-quality views and $X \in \{5,10,20,30\}$.

Minkowski pooling can also be applied to per-view distortion scores $d_v$, obtained by converting each quality score so that larger values indicate stronger perceived degradation:
\begin{equation}
D_{\mathrm{Minkowski}}
=
\left(
\frac{1}{N}
\sum_{v=1}^{N}
d_v^p
\right)^{1/p},
\end{equation}
where $p \in \{3,5,7,9\}$. These two strategies are used to assess whether subjective judgments are driven more by average quality or by localized severe degradations.

\subsection{Statistical Evaluation Criteria}
\label{sec:evaluation_criteria}

Objective metric reliability is assessed by comparing predicted scores with subjective scores using complementary statistical indicators for monotonicity and prediction accuracy. Spearman Rank-Order Correlation Coefficient (SRCC) measures ranking consistency, while Pearson Linear Correlation Coefficient (PLCC) and Root Mean Squared Error (RMSE) measure prediction accuracy after nonlinear mapping.

Because objective and subjective quality scores are typically nonlinearly related, a logistic mapping is applied before computing PLCC and RMSE. For an objective prediction $m$, the mapped prediction $\hat{s}(m)$ is
\begin{equation}
\hat{s}(m)
=
\beta_2
+
\frac{\beta_1-\beta_2}
{1+\exp\left(-\frac{m-\beta_3}{|\beta_4|}\right)},
\end{equation}
where $\beta_1,\ldots,\beta_4$ are fitted parameters. The mapping is fitted separately for each metric and evaluation subset.

\section{Benchmark Results}
\label{sec:benchmark_results}
% \textcolor{red}{Saeed will draft, Mylene to review} % reviewed! 
This section presents subjective and objective benchmark results from the proposed evaluation framework. We first analyze subjective scores, including their distribution, scene-wise behavior, and the impact of selective pairwise refinement. We then compare objective metrics with the final subjective ground truth, focusing on overall prediction accuracy, sensitivity to the pooling strategy, local agreement in ambiguous regions, and robustness across distortion families.

\subsection{Overall Subjective Assessment Results}
\label{sec:subjective_quality_Results}

\subsubsection{Distribution of Subjective Scores}
\label{sec:score_distribution}

\begin{figure}[t]
\centering
\includegraphics[width=\linewidth]{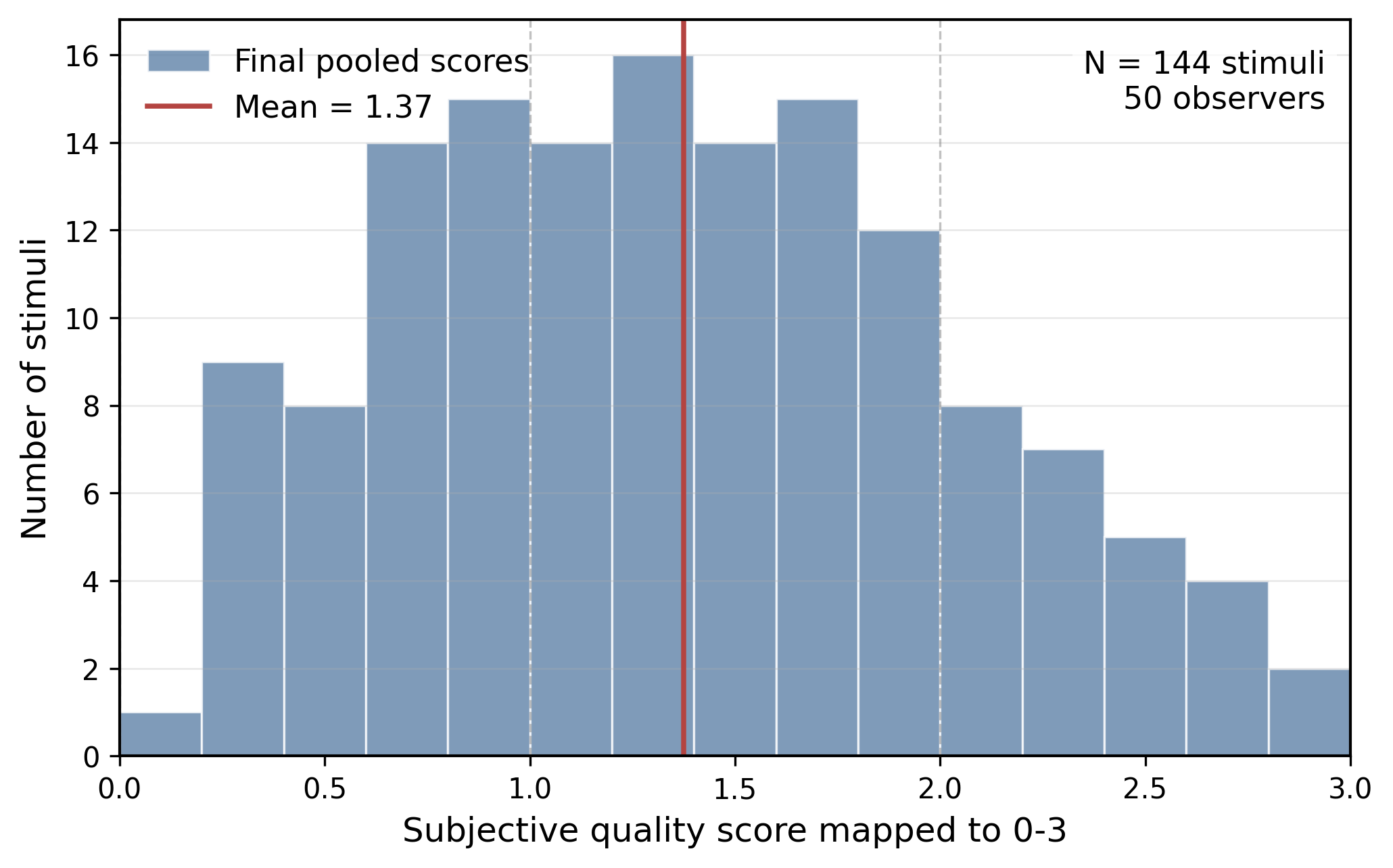}
\caption{Distribution of subjective scores of 144 stimuli. Scores are mapped to a 0--3 degradation scale, where lower values indicate better perceived quality.}
\label{fig:subjective_score_distribution}
\end{figure}

Figure~\ref{fig:subjective_score_distribution} shows the distribution of final subjective scores after pooling responses from all 50 valid observers across all stimuli. The scores are mapped to a common 0--3 degradation scale, where lower values indicate higher perceived quality and 0 corresponds to reference quality. The scores span a wide perceptual range, from near-reference to strongly degraded conditions, with a denser distribution in the medium- to high-quality range, enabling finer discrimination there. This range is especially relevant for objective metric evaluation, as practical coding systems typically operate where artifacts are visible but not severe. It is also a challenging region for subjective assessment, where rating-based methods are more susceptible to saturation. The resulting score distribution thus supports evaluating both metric robustness to strong degradations and sensitivity to subtle perceptual differences in practical quality ranges.

\subsubsection{Scene-wise Subjective Quality Scores}
\label{sec}

\begin{figure*}[t]
\centering
\includegraphics[width=\linewidth]{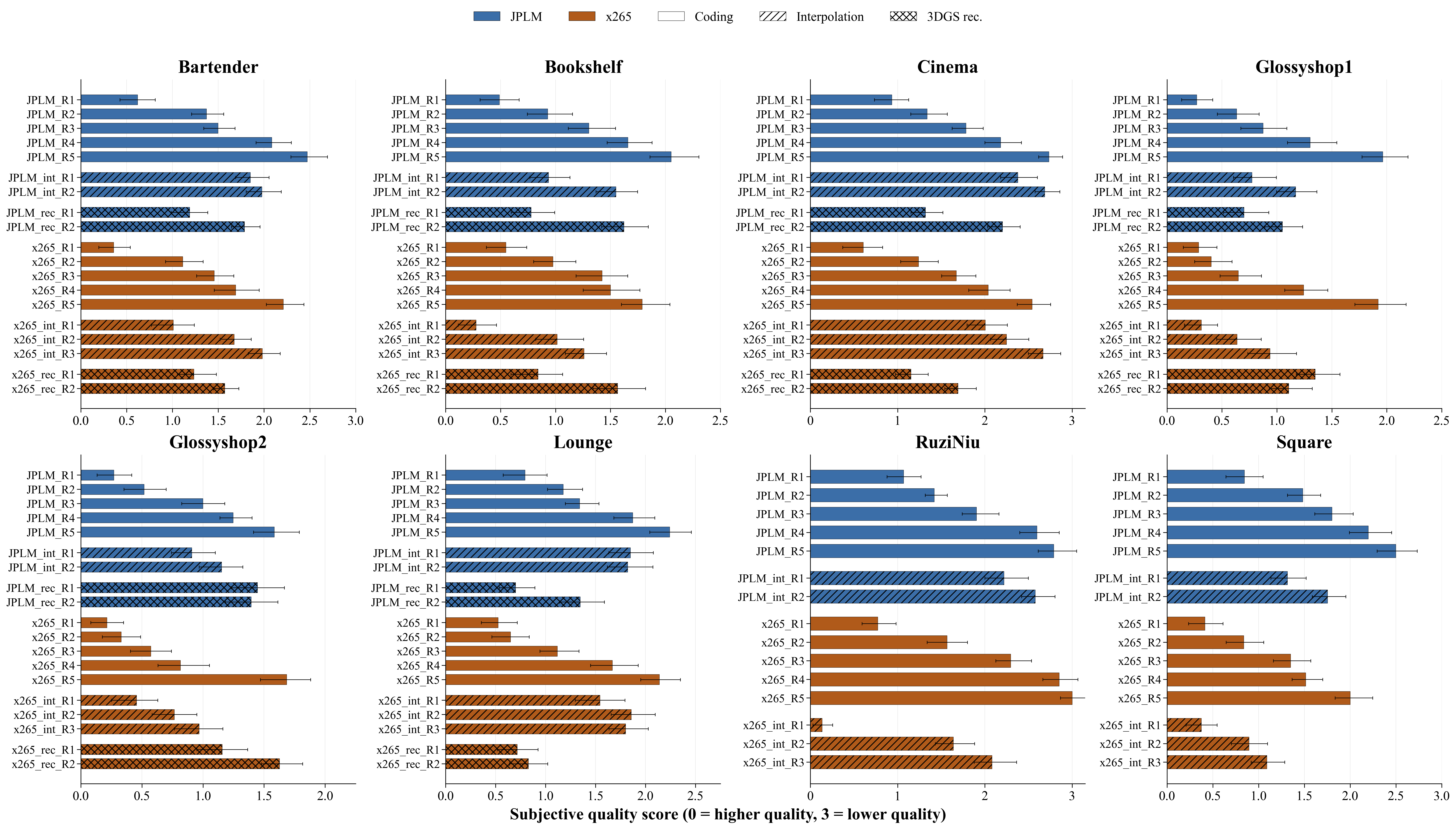}
\caption{Scene-wise subjective quality scores with confidence intervals.}
\label{fig:scene_wise_subjective_scores}
\end{figure*}

Figure~\ref{fig:scene_wise_subjective_scores} shows scene-wise subjective scores for coding-only, interpolation-based, and 3DGS-based reconstruction stimuli. The confidence intervals reflect the uncertainty of the pooled scores after integrating valid observer responses.
For coding-only stimuli, the chosen bitrate points yield a smooth quality progression within each scene, providing sufficiently fine steps to assess metric sensitivity to small perceptual changes. The interpolation- and 3DGS-based stimuli extend the benchmark by introducing artifacts that differ from conventional coding distortions. The results also reveal clear content dependency as the same processing family does not consistently produce the same subjective degradation across scenes, indicating that quality depends on scene spatio-angular properties.

\subsection{Validation of the Hybrid Subjective Scores}
\label{sec:pc_refinement_effect}

\subsubsection{Cross-Cohort Reproducibility and Consistency}
\label{sec:cross_cohort_consistency}

A standardized subjective procedure should yield consistent perceptual data across observer cohorts under the same protocol. To test this, we examined whether hybrid scores from one cohort could predict pairwise responses from another. Here, the primary cohort (Lab1) is the target set of pairwise observations, and the second cohort (Lab2) serves as an independent prediction source.

The analysis uses the Thurstone Case~V log-likelihood framework introduced in (2). It is applied to pairwise responses from Lab1's 20 observers, totaling 3,545 individual PC trials collected at the observer level during the selective PC stage. For each PC trial, the two stimuli are assigned predicted quality scores from one of three subjective score sets: the Lab1's hybrid scores, Lab1's DMOS (DSCS-only) scores, or the Lab2's hybrid scores. The score difference is converted to a predicted preference probability using the Thurstone Case~V model, and the log-likelihood of the observed Lab1 response is computed. Summing over all trials yields the total log-likelihood; a higher (less negative) value indicates stronger agreement between the predicted perceptual ordering and the observed pairwise choices.

Table~\ref{tab:repro_ll} summarizes the analysis. The highest likelihood occurs when Lab1's hybrid scores predict its own PC responses, indicating strong within-cohort consistency. Using hybrid scores from Lab2 lowers the likelihood because they come from an independent cohort, but this drop is much smaller than for the DSCS-only DMOS baseline, showing that the hybrid scores capture more cohort-consistent pairwise perceptual structure than the rating-only baseline.

\begin{table}[t]
\centering
\caption{Cross-cohort log-likelihood (LL) evaluation on 3,545 pairs from the first cohort (Lab1). Higher LL indicates better agreement with the observed pairwise choices.}
\label{tab:repro_ll}
\begin{tabular}{lccc}
\toprule
\textbf{Method} & \textbf{Total LL} & \textbf{Avg. LL / pair} & $\boldsymbol{\Delta}$\textbf{LL} \\
\midrule
Lab~1 predicts itself & $-3608.24$ & $-1.018$ & $0$ \\
Lab~2 predicts Lab~1  & $-3750.31$ & $-1.058$ & $-142.07$ \\
DMOS predicts Lab~1   & $-3967.89$ & $-1.119$ & $-359.64$ \\
\bottomrule
\end{tabular}
\end{table}

\begin{figure}[t]
\centering

\includegraphics[width=0.95\linewidth]{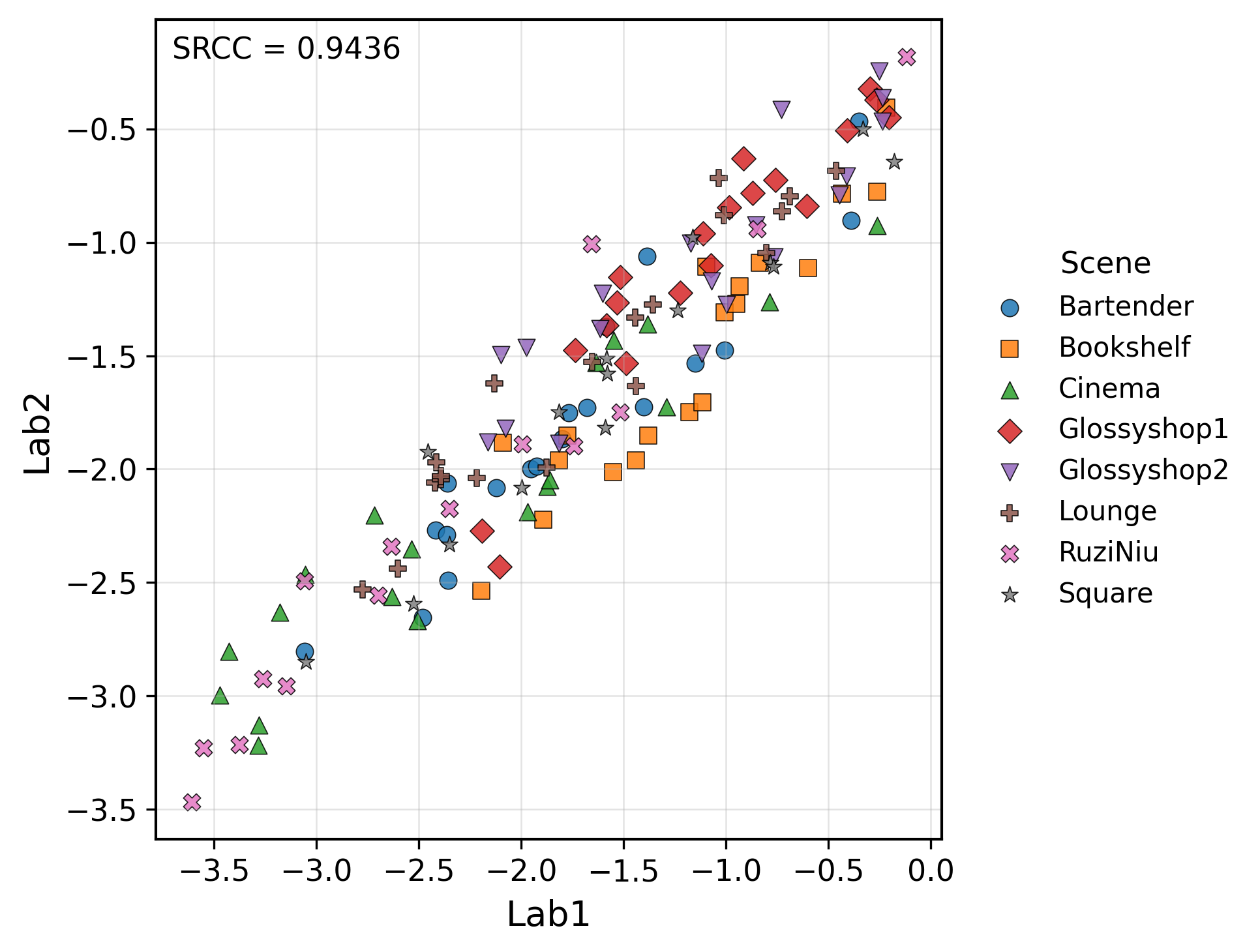}
\caption{Cross-laboratory consistency of final subjective scores. Each point corresponds to one benchmark stimulus, and colors/markers indicate different scenes.}
\label{fig:cross_lab_scatter}
\end{figure}

As a complementary score-level validation, Fig.~\ref{fig:cross_lab_scatter} compares the final hybrid scores from Lab1 and Lab2. The two sets show strong monotonic agreement, with an SRCC of 0.943 across the benchmark stimuli, indicating highly consistent perceptual rankings. Together with the pairwise log-likelihood analysis, this supports the cross-cohort reliability and reproducibility of the hybrid subjective approach.

\subsubsection{Effectiveness of the Selective Pairwise Refinement}
\label{sec:pc_selective_effect}

The selective pairwise comparison stage refines ambiguous regions of the DSCS scale while preserving the global reference-anchored structure from the first-stage ratings. DSCS-only and hybrid scores show strong agreement (SRCC = 0.984 over 144 benchmark stimuli), confirming that the hybrid reconstruction maintains the DSCS-derived quality structure. This stability in global ranking is desirable, as the PC stage is meant to make local adjustments without altering the reference-based structure or interpretability of the DSCS scale.

To assess the added value of the selective PC stage, we focus on PC trials where the rating stage has low resolution: same-category comparisons within \emph{Slightly worse} ($-1$) and \emph{Worse} ($-2$) DSCS levels, which were targeted for refinement. We test whether the hybrid reconstruction explains these locally ambiguous pairwise judgments better than DSCS-only scores.

We conducted a leave-one-observer-out evaluation on the integrated observer set ($N=50$). For each observer, we estimated DSCS-only and hybrid DSCS+PC scores from the other 49 observers, then evaluated these scores on the held-out observer’s PC responses. As in the previous log-likelihood analysis, for each held-out pairwise trial between stimuli $i$ and $j$, we converted the score difference to a predicted preference probability via the Thurstone Case~V model and computed the log-likelihood of the observer’s choice under this probability. This yielded one likelihood for DSCS-only and one for the hybrid scores; their difference indicates whether the hybrid refinement better predicts held-out pairwise judgments than the DSCS-only scale.

Table~\ref{tab:hybrid_advantage_loo} reports leave-one-observer-out log-likelihoods over 9,378 ambiguous pairwise trials from 50 observers. Hybrid DSCS+PC scores yield a much higher likelihood, with a total log-likelihood gain of 572.92 and an average gain of 0.0611 per trial. This indicates that the selective PC stage captures perceptual information that is not present in DSCS-only scores, even when the tested observer is excluded from the score estimation. The likelihood gain was positive for all eight scenes, which shows that the refinement effect is not driven by a single content.

\begin{table}[t]
\centering
\caption{Leave-one-observer-out evaluation log-likelihoods (LL) of ambiguous pairwise trials. Higher LL indicates better agreement with the observed pairwise choices. The hybrid DSCS+PC model improves over the DSCS-only baseline by $\Delta$LL=$+572.92$.}
\label{tab:hybrid_advantage_loo}
\begin{tabular}{lcc}
\toprule
\textbf{Score model} & \textbf{Total LL} & \textbf{Avg. LL/trial} \\
\midrule
DSCS-only & $-10534.19$ & $-1.1233$ \\
Hybrid DSCS+PC & $-9961.26$ & $-1.0622$ \\
\bottomrule
\end{tabular}
\end{table}

\subsection{Objective Metrics Analysis}

\subsubsection{Overall Performance}
\label{sec:overall_metric_performance}
Table~\ref{tab:metric_performance_mean_pooling_coding_full} reports the performance of objective metrics on the coding-only subset and the full benchmark in terms of SRCC, PLCC, and RMSE, using the final hybrid subjective scores as ground truth. Objective metrics perform substantially better on the coding-only subset, indicating that conventional coding distortions are easier to predict than the heterogeneous distortions from interpolation and 3DGS reconstruction. Among the evaluated metrics, ST-LPIPS, CVVDP, and IW-SSIM perform best overall. Deep learning-based metrics such as LPIPS and DISTS are also competitive, suggesting that deep feature-space distances capture perceptual degradations not fully represented by traditional signal-fidelity measures. In contrast, classical fidelity metrics such as PSNR, SSIM, and FSIM agree less with subjective scores, as interpolation and 3DGS-based reconstruction can introduce geometric shifts, view-dependent inconsistencies, and local pixel misalignment that strongly penalize pixel-based metrics, while feature-based and shift-tolerant approaches like ST-LPIPS remain more consistent.

\begin{table}[t]
\centering
\caption{Metrics performance on the coding-only subset and on the full benchmark in terms of SRCC, PLCC, and RMSE.}
\label{tab:metric_performance_mean_pooling_coding_full}
\tiny
\setlength{\tabcolsep}{1.5pt}
\resizebox{0.95\columnwidth}{!}{%
\begin{tabular}{lcccccc}
\toprule
\multirow{2}{*}{\textbf{Metric}} 
& \multicolumn{3}{c}{\textbf{Coding-only} ($N=80$)} 
& \multicolumn{3}{c}{\textbf{Full} ($N=144$)} \\
\cmidrule(lr){2-4}
\cmidrule(lr){5-7}
& \textbf{SRCC} $\uparrow$ & \textbf{PLCC} $\uparrow$ & \textbf{RMSE} $\downarrow$
& \textbf{SRCC} $\uparrow$ & \textbf{PLCC} $\uparrow$ & \textbf{RMSE} $\downarrow$ \\
\midrule
ST-LPIPS & 0.889 & 0.903 & 0.353 & \textbf{0.776} & \textbf{0.812} & \textbf{0.448} \\
CVVDP & \textbf{0.896} & \textbf{0.908} & \textbf{0.346} & 0.756 & 0.768 & 0.491 \\
IW-SSIM & 0.896 & 0.907 & 0.347 & 0.739 & 0.755 & 0.502 \\
LPIPS & 0.826 & 0.843 & 0.444 & 0.727 & 0.765 & 0.489 \\
MAD & 0.857 & 0.871 & 0.405 & 0.702 & 0.729 & 0.525 \\
DeepDC & 0.757 & 0.761 & 0.535 & 0.690 & 0.702 & 0.546 \\
GMSD & 0.873 & 0.884 & 0.386 & 0.686 & 0.698 & 0.549 \\
DISTS & 0.755 & 0.761 & 0.535 & 0.684 & 0.689 & 0.556 \\
MS-SSIM & 0.841 & 0.861 & 0.418 & 0.675 & 0.698 & 0.549 \\
VMAF & 0.785 & 0.764 & 0.532 & 0.625 & 0.615 & 0.605 \\
NLPD & 0.794 & 0.810 & 0.484 & 0.623 & 0.647 & 0.584 \\
VIF & 0.788 & 0.793 & 0.462 & 0.662 & 0.661 & 0.550 \\
PSNR & 0.713 & 0.750 & 0.545 & 0.553 & 0.601 & 0.613 \\
IV-PSNR & 0.620 & 0.675 & 0.608 & 0.531 & 0.601 & 0.613 \\
SSIM & 0.650 & 0.699 & 0.589 & 0.513 & 0.558 & 0.636 \\
FSIM & 0.652 & 0.659 & 0.620 & 0.489 & 0.505 & 0.662 \\
IV-SSIM & 0.496 & 0.514 & 0.707 & 0.406 & 0.487 & 0.670 \\
\bottomrule
\end{tabular}%
}
\end{table}
%%%%%%NEW

\subsubsection{Objective Metric Resolving Power}
\label{sec:metric_local_agreement}

Beyond global performance, we analyze metric behavior in locally ambiguous regions of the subjective scale. This relates to objective-metric resolving power, i.e., the ability to detect small quality differences between perceptually similar stimuli. We assess how consistently objective metric differences match local subjective score differences for stimulus pairs ambiguous under DSCS. Besides characterizing objective metrics, this shows whether hybrid DSCS+PC scores provide a more finely resolved local subjective target for metric evaluation.

Let $\mathcal{A}$ be the set of unique ambiguous stimulus pairs. In total, 1,150 unique within-scene pairs were identified. A pair $(i,j)$ is in $\mathcal{A}$ if at least one observer, in Stage I, placed both stimuli in the same DSCS category, \emph{Slightly worse} ($-1$) or \emph{Worse} ($-2$), and the pair was then shown in the selective PC stage, where a PC judgment was collected. 
For each stimulus $i$, let $d_i$ be the pooled DSCS-only degradation score, $h_i$ the final hybrid score, and $m_i$ the score produced by a given objective metric. All scores are converted to a common degradation-oriented scale, denoted by $\tilde{d}_i$, $\tilde{h}_i$, and $\tilde{m}_i$, where larger values indicate stronger degradation. For each ambiguous pair $(i,j)\in\mathcal{A}$, we computed signed pairwise differences for the DSCS-only, hybrid, and objective metric scores, denoted as $\Delta \tilde{d}_{ij}$, $\Delta \tilde{h}_{ij}$, and $\Delta \tilde{m}_{ij}$, respectively. The agreement between objective metric differences and local subjective differences is then measured with SRCC:
\begin{equation}
\rho_{\mathrm{DSCS}}
=
\operatorname{SRCC}
\left(
\{\Delta \tilde{m}_{ij}\}_{(i,j)\in\mathcal{A}},
\{\Delta \tilde{d}_{ij}\}_{(i,j)\in\mathcal{A}}
\right),
\end{equation}
\begin{equation}
\rho_{\mathrm{Hybrid}}
=
\operatorname{SRCC}
\left(
\{\Delta \tilde{m}_{ij}\}_{(i,j)\in\mathcal{A}},
\{\Delta \tilde{h}_{ij}\}_{(i,j)\in\mathcal{A}}
\right).
\end{equation}
The gain reported is $\Delta \rho =\rho_{\mathrm{Hybrid}} - \rho_{\mathrm{DSCS}}$.
A positive $\Delta\rho$ means that the objective metric differences align better with the hybrid DSCS+PC ordering than with the DSCS-only ordering.   The metric-difference analysis shows modest but consistent gains (Table~\ref{tab:metric_local_agreement}), indicating that the refined hybrid target provides clearer local ordering in ambiguous regions and better assesses metric sensitivity to small perceptual differences than the DSCS-only target. These gains show that the hybrid target improves local subjective resolution over DSCS-only scores, enabling more informative evaluation of metric sensitivity to small perceptual differences.

\begin{table}[t]
\centering
\caption{Local subjective resolution and metric agreement in ambiguous regions for the full benchmark. The analysis is computed over 1150 ambiguous pairs.}% ``Unique $\Delta$'' denotes the number of unique rounded pairwise differences, i.e., $|\operatorname{unique}(\operatorname{round}(\Delta,3))|$.}
\label{tab:metric_local_agreement}
\small
\setlength{\tabcolsep}{3.8pt}
\renewcommand{\arraystretch}{0.9}
\begin{tabular}{@{}lccc@{}}
\toprule
\textbf{Target / metric} 
& \textbf{DSCS} 
& \textbf{Hybrid} 
& \textbf{Change} \\
\midrule
%Unique $\Delta$ values & 188 & 928 & \cellcolor{green!25}$+740$ \\
%Zero $\Delta$ values   & 13  & 0   & \cellcolor{green!15}$-13$ \\
\midrule
\textbf{Metric} 
& $\boldsymbol{\rho_{\mathrm{DSCS}}}$ 
& $\boldsymbol{\rho_{\mathrm{Hybrid}}}$ 
& $\boldsymbol{\Delta\rho}$ \\
\midrule
ST-LPIPS & 0.719 & 0.737 & \cellcolor{green!18}$+0.018$ \\
SSIM     & 0.735 & 0.751 & \cellcolor{green!16}$+0.016$ \\
MS-SSIM  & 0.761 & 0.775 & \cellcolor{green!14}$+0.014$ \\
LPIPS    & 0.699 & 0.712 & \cellcolor{green!13}$+0.013$ \\
IV-SSIM  & 0.728 & 0.741 & \cellcolor{green!13}$+0.013$ \\
NLPD     & 0.733 & 0.745 & \cellcolor{green!12}$+0.012$ \\
VIF      & 0.762 & 0.774 & \cellcolor{green!12}$+0.012$ \\
GMSD     & 0.763 & 0.775 & \cellcolor{green!12}$+0.012$ \\
IV-PSNR  & 0.808 & 0.820 & \cellcolor{green!12}$+0.012$ \\
IW-SSIM  & 0.760 & 0.771 & \cellcolor{green!11}$+0.011$ \\
VMAF     & 0.663 & 0.674 & \cellcolor{green!11}$+0.011$ \\
PSNR     & 0.746 & 0.756 & \cellcolor{green!10}$+0.010$ \\
MAD      & 0.819 & 0.828 & \cellcolor{green!9}$+0.009$ \\
CVVDP    & 0.771 & 0.780 & \cellcolor{green!9}$+0.009$ \\
DISTS    & 0.645 & 0.653 & \cellcolor{green!8}$+0.008$ \\
DeepDC   & 0.742 & 0.747 & \cellcolor{green!5}$+0.005$ \\
FSIM     & 0.713 & 0.717 & \cellcolor{green!4}$+0.004$ \\
\midrule
\textbf{Mean gain} & -- & -- & \cellcolor{green!12}$\mathbf{+0.011}$ \\
%\textbf{Positive gains} & -- & -- & \cellcolor{green!15}$\mathbf{17/17}$ \\
\bottomrule
\end{tabular}
\end{table}

\subsubsection{Metric Robustness Across Distortion Families}
\label{sec:metric_robustness}

To better understand the performance drop when view synthesis artifacts are included, we analyze how objective metrics transfer across distortion families. For each metric, a four-parameter logistic mapping was fitted using only coding-only stimuli, then applied to interpolation- and 3DGS-based stimuli. Thus, each reconstruction stimulus receives a predicted subjective score based solely on the coding-distortion relationship.

Let $\hat{h}_i^{\mathrm{cod}}$ be the predicted hybrid degradation score for reconstruction stimulus $i$ obtained by applying the coding-trained logistic mapping, and let $h_i$ be the corresponding hybrid subjective degradation score. The residual is defined as $r_i = h_i - \hat{h}_i^{\mathrm{cod}}$.
Its distribution shows whether a metric calibrated on coding artifacts transfers reliably to reconstruction artifacts. Positive residuals mean the coding-trained model underestimates reconstruction degradation; negative residuals mean it overestimates it.

\begin{figure}[t]
\centering
 \includegraphics[width=0.98\linewidth]{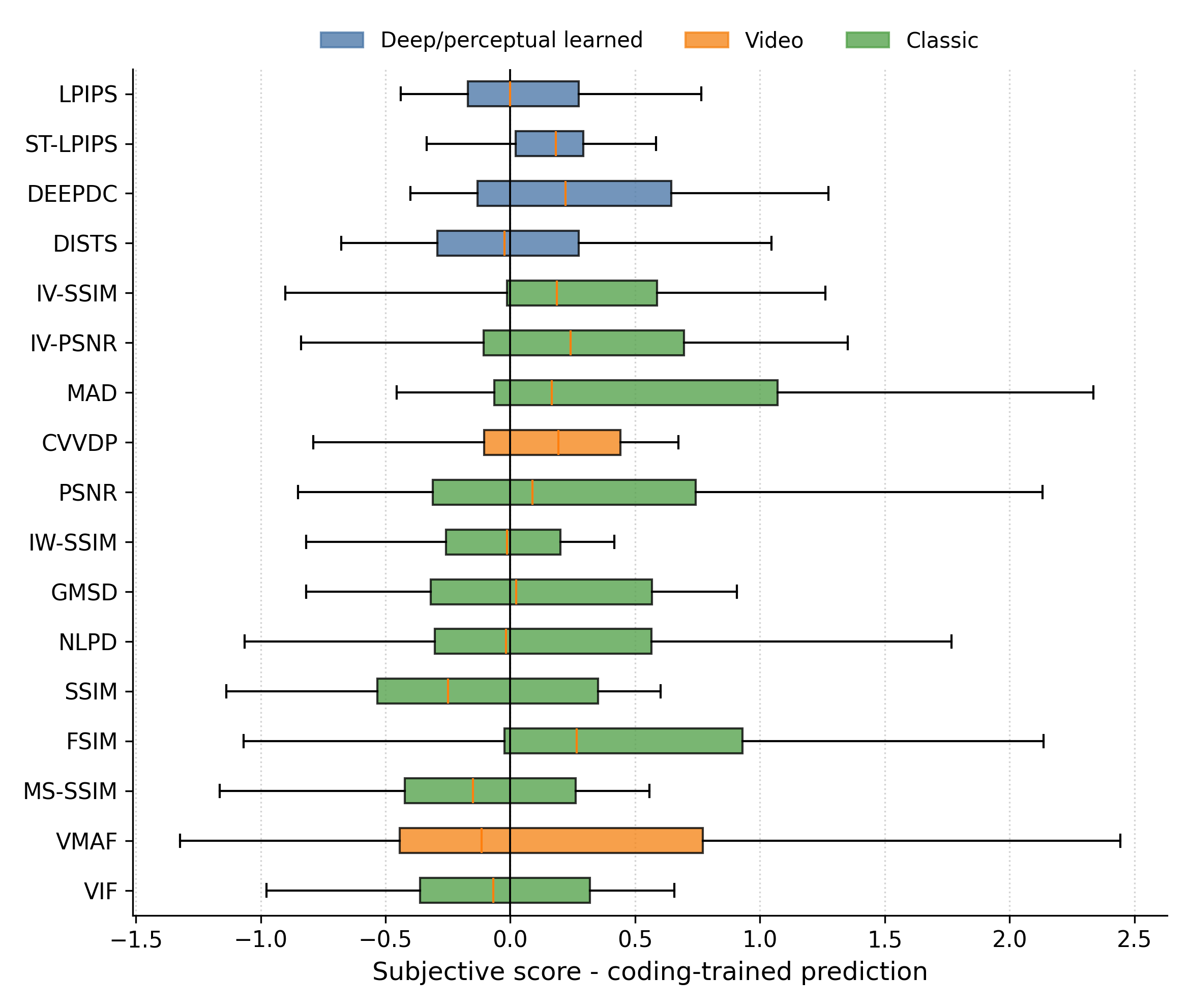}
\caption{Cross-distortion residual distribution. For each metric, a four-parameter logistic mapping was fitted using coding-only stimuli and then applied to reconstruction-based stimuli.}
\label{fig:reconstruction_residuals}
\end{figure}

Figure~\ref{fig:reconstruction_residuals} shows the residual distributions for reconstruction-based stimuli. Metrics that perform well on coding-only stimuli can still show systematic bias and large residual spread on reconstruction-based stimuli. For most top metrics, the median residual is positive, indicating that, after calibration on coding-only distortions, they tend to under-predict the perceived degradation of reconstruction-based stimuli. This suggests that view synthesis artifacts are not fully captured by metric behavior learned from coding-only conditions.

\subsubsection{Impact of Pooling}
\label{sec:pooling_strategy_effect}

Table~\ref{tab:metric_pooling_srcc} reports how view-pooling strategies affect image quality metric performance on the full benchmark. Mean pooling serves as the baseline, while Minkowski and worst-view pooling emphasize views with stronger predicted degradation. Pooling noticeably affects SRCC in a metric-dependent way. Learning-based metrics, including ST-LPIPS, LPIPS, DISTS, and DeepDC, generally improve with non-mean pooling. In particular, worst-view pooling gives clear gains for ST-LPIPS, LPIPS, and DISTS, suggesting that perceived quality is influenced not only by average view quality but also by localized poor views caused by view-dependent artifacts. In contrast, traditional fidelity metrics such as IW-SSIM, MS-SSIM, PSNR, and VIF are often best or near-best with mean pooling, and their performance can drop by other pooling methods. The results show that view pooling is important for LF metric evaluation and should be considered in future LF-specific metric design, especially for learning-based perceptual metrics.

\begin{table*}[t]
\centering
\caption{Effect of pooling strategy on SRCC for the full benchmark ($N=144$). Green cells indicate higher SRCC than mean pooling, while red cells indicate lower SRCC.}
\label{tab:metric_pooling_srcc}
\scriptsize
\setlength{\tabcolsep}{3pt}
\resizebox{0.85\textwidth}{!}{%
\begin{tabular}{lc|cccccccc}
\toprule
\textbf{Metric} 
& \textbf{Mean} 
& \textbf{Mink.-3} 
& \textbf{Mink.-5} 
& \textbf{Mink.-7} 
& \textbf{Mink.-9} 
& \textbf{Worst 5\%} 
& \textbf{Worst 10\%} 
& \textbf{Worst 20\%} 
& \textbf{Worst 30\%} \\
\midrule
ST-LPIPS 
& 0.776 
& \cellcolor{green!8}0.783 
& \cellcolor{green!8}0.786 
& \cellcolor{green!8}0.784 
& \cellcolor{green!8}0.783 
& \cellcolor{green!12}0.795 
& \cellcolor{green!12}0.795 
& \cellcolor{green!12}0.798 
& \cellcolor{green!12}\textbf{0.798} \\

LPIPS 
& 0.727 
& \cellcolor{green!8}0.736 
& \cellcolor{green!8}0.739 
& \cellcolor{green!12}0.742 
& \cellcolor{green!12}0.745 
& \cellcolor{green!18}\textbf{0.766} 
& \cellcolor{green!18}0.763 
& \cellcolor{green!12}0.757 
& \cellcolor{green!12}0.752 \\

DISTS 
& 0.684 
& \cellcolor{green!8}0.691 
& \cellcolor{green!8}0.693 
& \cellcolor{green!8}0.695 
& \cellcolor{green!8}0.697 
& \cellcolor{green!12}0.708 
& \cellcolor{green!12}0.707 
& \cellcolor{green!12}\textbf{0.708} 
& \cellcolor{green!12}0.705 \\

DeepDC 
& 0.690 
& \cellcolor{green!8}0.697 
& \cellcolor{green!4}0.692 
& \cellcolor{red!8}0.684 
& \cellcolor{red!8}0.681 
& \cellcolor{green!4}0.695 
& \cellcolor{green!8}\textbf{0.697} 
& \cellcolor{green!4}0.695 
& \cellcolor{green!4}0.695 \\

SSIM 
& 0.513 
& \cellcolor{green!4}0.516 
& \cellcolor{green!8}0.518 
& \cellcolor{green!4}0.517 
& \cellcolor{green!4}0.516 
& \cellcolor{green!8}0.524 
& \cellcolor{green!12}0.530 
& \cellcolor{green!12}\textbf{0.532} 
& \cellcolor{green!12}0.531 \\

IW-SSIM 
& \textbf{0.739} 
& \textbf{0.739} 
& \cellcolor{red!4}0.739 
& \cellcolor{red!8}0.731 
& \cellcolor{red!12}0.714 
& \cellcolor{red!25}0.682 
& \cellcolor{red!18}0.694 
& \cellcolor{red!18}0.706 
& \cellcolor{red!12}0.712 \\

MS-SSIM 
& \textbf{0.675} 
& \cellcolor{red!4}0.674 
& \cellcolor{red!4}0.673 
& \cellcolor{red!4}0.673 
& \cellcolor{red!4}0.671 
& \cellcolor{red!18}0.634 
& \cellcolor{red!12}0.646 
& \cellcolor{red!12}0.657 
& \cellcolor{red!8}0.661 \\

VIF 
& 0.662 
& \cellcolor{green!4}\textbf{0.663} 
& \cellcolor{red!4}0.659 
& \cellcolor{red!8}0.650 
& \cellcolor{red!12}0.638 
& \cellcolor{red!12}0.645 
& \cellcolor{red!8}0.651 
& \cellcolor{red!8}0.654 
& \cellcolor{red!8}0.656 \\

PSNR 
& \textbf{0.553} 
& \cellcolor{red!8}0.540 
& \cellcolor{red!12}0.535 
& \cellcolor{red!12}0.527 
& \cellcolor{red!18}0.510 
& \cellcolor{red!25}0.476 
& \cellcolor{red!25}0.486 
& \cellcolor{red!25}0.501 
& \cellcolor{red!18}0.513 \\

\bottomrule
\end{tabular}%
}
\end{table*}

\section{Discussion and Implications for Objective Metric Design}
\label{sec:disc}

The results validate the hybrid DSCS+PC method for objective metric evaluation. Rating methods like DSCS provide a stable, interpretable quality scale but lose sensitivity when stimuli are perceptually close. By adding selective pairwise comparisons in these ambiguous regions, the hybrid method produces a finer-grained subjective target for testing metrics under small perceptual differences, while preserving the DSCS reference-based structure.

The benchmark reveals clear limitations of current objective metrics under LF processing pipelines based on interpolation and view reconstruction. While several metrics match subjective scores for coding-only stimuli, their performance degrades for interpolation- and 3DGS-based reconstructions, which introduce non-traditional artifacts such as view-dependent inconsistencies, geometric shifts, local misalignments, and rendering distortions. As a result, metrics effective for compression do not reliably generalize to reconstruction-based scenarios, a crucial issue for future LF systems that will combine compression and view reconstruction.

Results show that robust objective LF metrics must capture more than spatial fidelity in single views. Classical pixel-based metrics (PSNR, SSIM, etc.) correlate poorly with subjective scores over the full benchmark and fail to reflect perceptual tolerance to local geometric shifts or angular inconsistencies. Feature-based metrics (ST-LPIPS, CVVDP, IW-SSIM, LPIPS) agree better with subjective ground truth, indicating that effective LF quality prediction requires feature-level similarity modeling and mechanisms aligned with human visual sensitivity to distortions.

The pooling analysis further shows that stimulus-level LF quality is not always well represented by simple mean aggregation. For several perceptual metrics, weighting lower-quality views more heavily improves correlation with subjective scores, indicating that observers are strongly influenced by localized poor views, especially when only some views contain severe view-dependent artifacts. Therefore, a reliable LF metric should accurately estimate per-view quality and model how quality variations across views determine overall perceived quality.

These findings underscore the need for standardized objective metrics for LF content. As LF coding and representation move beyond traditional coding pipelines, metrics must be evaluated under both compression and reconstruction artifacts. A standardized framework is therefore essential to define subjective methods, stimulus generation, and statistical criteria in a transparent, reproducible way. The benchmark and analysis in this work advance such a framework by providing shared subjective ground truth, diverse distortions, and a structured protocol for evaluating objective metric performance.

% This paper advances standardized light field quality assessment by unifying benchmark design, hybrid subjective evaluation, and objective metric analysis in a single framework. The hybrid DSCS+PC method selectively applies pairwise comparisons to refine ambiguous regions of a DSCS rating scale, preserving the efficiency and interpretability of ratings while adding local perceptual precision without the complexity of exhaustive pairwise testing. Beyond metric evaluation, this framework offers a practical way to obtain more reliable quality scores. Benchmark results show that current objective metrics remain limited for reconstruction- and interpolation-based processing, even when they work well for coding-only distortions. The analysis also highlights the importance of distortion-family robustness, view-pooling strategy, and consistency under small perceptual differences. Overall, this work provides a reproducible subjective reference, diverse stimuli, and a structured protocol for more reliable development, comparison, and standardization of light field quality metrics.

\section{Conclusion}
\label{sec:conc}
This paper presents a framework for light field QA that unifies benchmark generation, hybrid subjective evaluation, and anchor objective metric analysis. The hybrid DSCS+PC method selectively uses pairwise comparison only to refine ambiguous parts of a reference-based DSCS scale. This preserves the interpretability and efficiency of rating-based assessment while adding perceptual detail where ratings are less discriminative, without the high complexity of exhaustive pairwise comparison. Benchmark results show that current objective metrics remain limited for reconstruction- and interpolation-based processing, even when they perform well for coding-only distortions. The analysis further shows that metric evaluation should consider distortion-family robustness, view-pooling strategy, and local resolving power. Overall, the framework offers a reproducible subjective target and evaluation protocol to support more reliable development, comparison, and standardization of objective light field quality metrics.

% or


%\section*{Acknowledgment}


\ifCLASSOPTIONcaptionsoff
  \newpage
\fi



%\bibliographystyle{IEEEtran}
\bibliographystyle{IEEEtran}
\bibliography{references_up}


%\vfill


\end{document}